\documentclass[10pt, conference]{IEEEtran}
\IEEEoverridecommandlockouts
\usepackage{amsmath, amssymb, amsfonts}
\usepackage{algorithmic}
\usepackage{cite}
\usepackage{graphicx}
\usepackage{quantikz}
\usepackage{booktabs}
\usepackage{multirow}
\usepackage{textcomp}
\usepackage{tikz}
\usepackage{xcolor}
\usepackage{hyperref}
\usetikzlibrary{positioning, calc, backgrounds, arrows.meta}
\def\BibTeX{{\rm B\kern-.05em{\sc i\kern-.025em b}\kern-.08em
    T\kern-.1667em\lower.7ex\hbox{E}\kern-.125emX}}
\begin{document}

\title{QuantumBoostNet: Hybrid Classical-Quantum Cardiac View Identification}

\author{\IEEEauthorblockN{Mihai Udrescu-Milosav}
\IEEEauthorblockA{\textit{Department of Computers and Information Technology} \\
\textit{Politehnica University Timi\c{s}oara}\\
Timi\c{s}oara, Romania \\
mihai-alexandru.udrescu-milosav@student.upt.ro}
\and
\IEEEauthorblockN{\c Stefan-Alexandru Jura}
\IEEEauthorblockA{\textit{Department of Computers and Information Technology} \\
\textit{Politehnica University Timi\c{s}oara}\\
Timi\c soara, Romania \\
stefan.jura@student.upt.ro}
\and
\IEEEauthorblockN{Mihai Udrescu}
\IEEEauthorblockA{\textit{Department of Computers and Information Technology} \\
\textit{Politehnica University Timi\c{s}oara}\\
Timi\c soara, Romania \\
mihai.udrescu@cs.upt.ro}
\and
\IEEEauthorblockN{Gerhard-Paul Diller}
\IEEEauthorblockA{\textit{Department of Cardiology \& Angiology III} \\ \textit{Adult Congenital \& Valvular Heart Disease Center} \\
\textit{University Hospital M\"uenster}\\
M\"unster, Germany \\
gerhard.diller@ukmuenster.de}}

\maketitle

\begin{abstract}
Accurate identification of the correct view or angle in cardiac ultrasound (echocardiogram) is critical for cardiologic imaging, precise anatomical interpretation, and reducing clinical errors. Most state-of-the-art classical models perform well on standard benchmarks but give suboptimal results in specialized medical imaging due to high noise levels. To address these challenges, this work proposes the hybrid classical-quantum architecture QuantumBoostNet, which combines a classical backbone with two heads: one classical and one quantum, a parametrized 10-qubit quantum circuit. The main contribution of this work is training in two stages, with an adaptive transition between heads controlled by a mixing parameter that monitors loss dynamics. Extensive experiments show that QuantumBoostNet outperforms the implemented baselines under matched training conditions. Statistically significant gains appear on FashionMNIST ($p_t=9.91\!\times\!10^{-6}$) and MNIST ($p_t=1.29\!\times\!10^{-5}$), with a less significant improvement on the echocardiography task ($p_t=0.0711$). The model also shows robustness to noise. These findings support continued development of hybrid classical-quantum models for specialized medical imaging applications.

\end{abstract}



\section{Introduction}

Accurate view identification is critical for meaningful echocardiographic interpretation because each standard imaging plane reveals a distinct subset of cardiac anatomy and enables specific quantitative measurements. Off-axis or misidentified views cause anatomical misinterpretation and measurement error \cite{mitchell2019guidelines}\cite{lang2015recommendations}. In machine learning pipelines, view identification acts as the routing step for downstream segmentation, chamber quantification, and disease detection, not just a preprocessing step \cite{madani2018}\cite{zhang2018fully}. Recent multi-task systems also combine view classification with image-quality assessment \cite{li2024multitask}.

View identification in ultrasound imaging is challenging due to speckle, low contrast, blurred tissue boundaries, attenuation, and operator-dependent variability in echocardiograms. These factors impair both human interpretation and deep-learning feature extraction \cite{li2024multitask}. Such limitations are especially harmful for tasks requiring precise border localization or temporal tracking, like LV segmentation, Simpson-based ejection-fraction estimation, and strain analysis \cite{lang2015recommendations}\cite{leitman2022pitfalls}. Image quality degradation from reduced frame rate or lossy compression can remove motion information essential for speckle-tracking analysis and may cause misdiagnosis \cite{leitman2022pitfalls}. Moreover, performance in general cardiology cohorts does not always apply to specialized clinical settings. For example, in congenital or structural heart disease cases, a generic deep-learning view classifier showed significantly reduced accuracy, while a disease-specific model performed better. These findings highlight the need for noise-robust, population-specific training data and models in medical computer vision\cite{li2024multitask}\cite{wegner2022accuracy}. 

Most state-of-the-art classical models achieve high performance in standard view identification on curated ultrasound images \cite{madani2018} but often produce suboptimal results in specialized medical imaging tasks, such as view identification in ultrasound images from patients with congenital or structural heart disease (C/SHD) \cite{wegner2022accuracy}.
This paper investigates whether hybrid classical-quantum machine learning methods can improve view identification accuracy in C/SHD ultrasound images and enhance image classification. It also examines whether the training method of hybrid heads matters as much as the quantum circuit itself.
Accordingly, the contributions are:

\begin{itemize}
\item A comprehensive analysis of view identification accuracy in C/SHD ultrasound images on a $210,345$-frame dataset using classical and hybrid classical-quantum methods.

\item The introduction of a novel two-phase training protocol for hybrid classical-quantum models, phase-switched optimization of the quantum and classical heads, instantiated in \emph{QuantumBoostNet}, an architecture integrating a classical backbone with two parallel heads: one classical and one quantum. \emph{QuantumBoostNet} achieves the highest mean accuracy among all of the implemented classical and hybrid models in view identification for C/SHD ultrasound images.

\item An evaluation showing that \emph{QuantumBoostNet} achieves statistically significant accuracy improvements over the implemented baselines on FashionMNIST and MNIST.
\end{itemize}



\section{Background}

\subsection{Classical state-of-the-art models}


Madani et al.\cite{madani2018} pioneered echocardiographic view classification with a convolutional neural network (CNN) that distinguishes 15 views from 267 transthoracic echocardiograms. Their model achieved 97.8\% accuracy on video clips and 91.7\% on still images, surpassing the performance of board-certified echocardiographers. Kusunose et al.\cite{kusunose2020} confirmed 98.1\% accuracy on an independent cohort of five views using 5-fold cross-validation, establishing the clinical feasibility of CNN-based classifiers. Li et al.\cite{li2024multitask} introduced a multi-task model that performs view classification and image quality assessment across 6 standard views and an “others” category, achieving 97.8\% accuracy on 170,311 images. These studies show that CNNs can achieve near-expert performance.



In echocardiographic analysis, CNN backbones are commonly used as feature extractors. The EchoNet pipeline by Ghorbani et al.~\cite{ghorbani2020deep} used an Inception-ResNet-v1-based architecture instead of ResNet-18. Note that recent high-performing systems such as EchoFM \cite{echofm} and EchoViewCLIP \cite{echoviewclip} operate on video clips and exploit temporal context across frames classify individual static frames.

\subsection{Hybrid classical-quantum models}


A qubit is the quantum analog of a bit, a superposition of the basis states $|0\rangle$ and $|1\rangle$;  an $N$-qubit register occupies a $2^N$-dimensional state space. Quantum gates are unitary transformations of this state, and information is extracted through measurement, which yields expectation values of observables such as the Pauli-$Z$ operator, real numbers in $[-1,1]$ that serve as the circuit's outputs. Because simulating an $N$-qubit register on classical hardware requires memory exponential in $N$, practical hybrid models operate with small registers (in our cases, $N$ is $4$ or $10$).
Parameterized quantum circuits (PQCs) have become a useful component in machine learning~\cite{benedetti2019pqc, schuld2020circuit}, with applications in computer vision~\cite{meli2025quantum}. PQCs encode classical data into quantum states using angle or amplitude embedding; they implement sequences of trainable unitary gates, such as single-qubit rotations combined with entangling CNOT gates, and produce predictions through Pauli measurements. Trainable parameter optimization uses the parameter-shift rule~\cite{mitarai2018, schuld2019analytic}, which yields exact analytic gradients, and integrates with automatic differentiation frameworks~\cite{bergholm2022pennylane}.


The simplest hybrid paradigm is the sequential architecture: a classical feature extractor is truncated and followed by a quantum component that produces the final class logits~\cite{mari2020transfer}. Mari et al.~\cite{mari2020transfer} demonstrated this classical-to-quantum transfer learning method by replacing the final layer of a pretrained ResNet-18 with a dressed quantum circuit. Henderson et al.~\cite{henderson2020quanvolution} proposed a paradigm where random quantum circuits act as convolutional filters, called quanvolutional layers, to preprocess image patches before classical layers.


Recent studies have refined hybrid classical-quantum representations. Alavi et al.~\cite{alavi2025fusion} regarded hybrid learning as a multimodal fusion problem. Anwar et al.~\cite{anwar2025multiclass} reached the same objective in the case of multiclass image classification by reusing the qubit states that were discarded during QCNN pooling. Zhang et al.~\cite{zhang2025readout} combined the quantum features with the raw inputs before carrying out the classification. Chaves et al.~\cite{chaves2026moe} showed that routing mechanisms can make hybrid models competitive.


Another direction advances transfer learning within hybrid classical-quantum frameworks. Kim, Huh, and Park~\cite{kim2023cqcnn} expanded this concept for quantum convolutional neural networks (QCNNs), demonstrating that compact quantum convolutional models can leverage features transferred from pre-trained classical convolutional neural networks (CNNs) under near-term circuit limitations. Martín-Pérez et al.~\cite{martinperez2026noisy} integrated variational quantum classifiers with frozen convolutional backbones. Hu et al.~\cite{hu2026tlqnn} argued that current classical-quantum transfer learning pipelines are underparameterized, and introduced amplitude-encoding-based models with multi-layer ansätze to increase the quantum parameter space. Yogaraj et al.~\cite{yogaraj2025pvcqtl} proposed post-variational classical-quantum transfer learning, substituting fully variational quantum heads with modified observables and related post-variational architectures.



\subsection{Limitations of hybrid classical-quantum models}


A significant limitation of sequential hybrid models is the information bottleneck: high-dimensional classical features must be compressed to fit the limited qubits on near-term quantum devices. To address this, Kordzanganeh et al. \cite{kordzanganeh2023phn} proposed Parallel Hybrid Networks (PHNs), where inputs are processed simultaneously by a variational quantum circuit and a classical multi-layer perceptron, and their outputs are combined using trainable weights. 
Also, the Lean Classical-Quantum Hybrid Neural Network (LCQHNN)~\cite{liu2025lcqhnn} combines a classical multi-channel CNN feature extractor with a streamlined four-layer variational quantum circuit and reports strong performance on FashionMNIST~\cite{xiao2017fashion}, MNIST~\cite{lecun1998gradient}, and a CIFAR-10~\cite{krizhevsky2009learning} subset while reducing circuit complexity.


\section{QuantumBoostNet}

\subsection{Model description}

\begin{figure*}[!t]
\centering
\makebox[\textwidth]{%
\scalebox{0.671}{%
\begin{tikzpicture}[
    >=Latex,
    node distance=1.2cm,
    block/.style={
        draw,
        fill=white,
        rectangle,
        minimum width=2.1cm,
        minimum height=4.0cm,
        align=center
    }
]

\node (img) {\includegraphics[height=7.5cm,keepaspectratio]{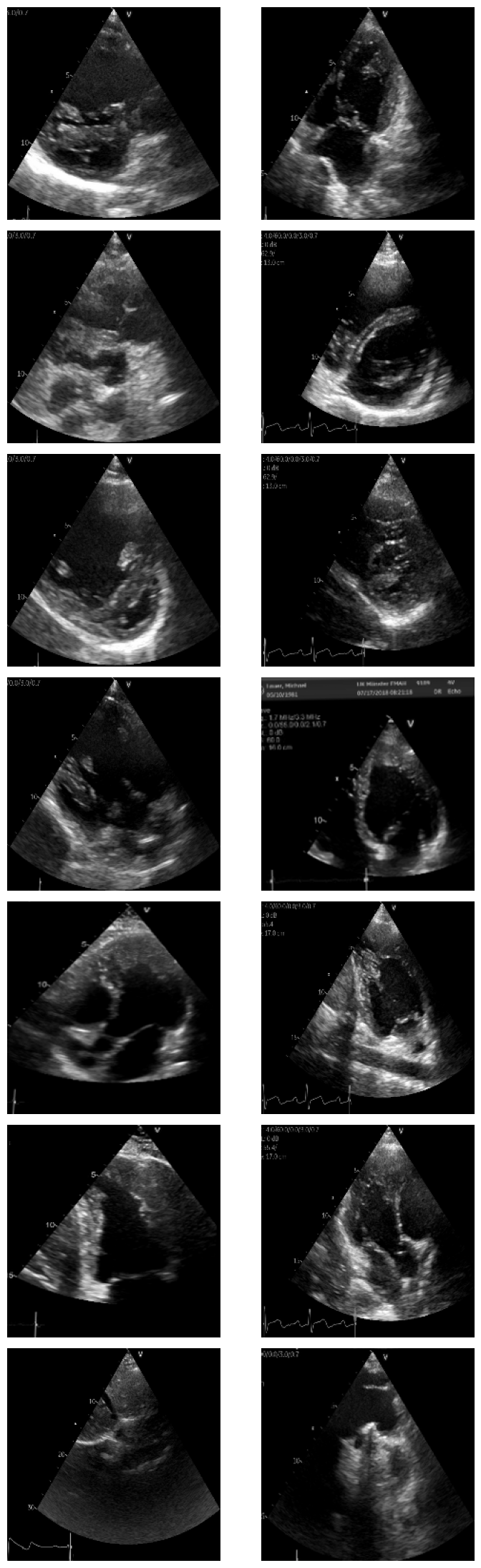}};

\draw[thick]
    ([xshift=2mm]img.north east) -- ([xshift=6mm]img.north east)
    ([xshift=6mm]img.north east) -- ([xshift=6mm]img.south east)
    ([xshift=6mm]img.south east) -- ([xshift=2mm]img.south east);

\node[below=2mm of img, align=center] {\small Noisy Cardiac\\Ultrasound Images};

\node[inner sep=0pt, right=of img] (resnet) {%
\begin{tikzpicture}[scale=0.75, baseline=(current bounding box.center)]

\def\W{3.0}
\def\H{3.0}
\def\s{0.25}
\def\J{1.5}

\draw[fill=blue!20] ({-2*\s},{ 2*\s}) rectangle ({\W-2*\s},{\H+2*\s});
\draw[fill=blue!20] ({-1*\s},{ 1*\s}) rectangle ({\W-1*\s},{\H+1*\s});
\draw[fill=blue!20] ({ 0*\s},{ 0*\s}) rectangle ({\W-0*\s},{\H+0*\s});

\node at (0.125,-0.5) {\Huge $\cdots$};

\draw[fill=blue!20] ({\J},{-\J}) rectangle ({\J+\W},{\H-\J});

\end{tikzpicture}%
};

\node[below=2mm of resnet, align=center] {\small ResNet Backbone};

\node[
    draw,
    fill=white,
    rectangle,
    inner sep=2mm,
    right=of resnet
] (qhead) {%
\resizebox{5.2cm}{!}{%
\begin{quantikz}[row sep={0.75cm,between origins}, column sep=0.25cm, thin lines]
\lstick{$q_{0}$} & \gate{R_x} & \gate[10,style={fill=violet!40}]{\shortstack{Strongly \\ Entangled \\ Layer 1}} & \gate[10,style={fill=violet!40}]{\shortstack{Strongly \\ Entangled \\ Layer 2}} & \gate[10,style={fill=violet!40}]{\shortstack{Strongly \\ Entangled \\ Layer 3}} & \gate[10,style={fill=violet!40}]{\shortstack{Strongly \\ Entangled \\ Layer 4}} & \meter{} \\
\lstick{$q_{1}$} & \gate{R_x} &                                                          &                                                          &                                                          &                                                          & \meter{} \\
\lstick{$q_{2}$} & \gate{R_x} &                                                          &                                                          &                                                          &                                                          & \meter{} \\
\lstick{$q_{3}$} & \gate{R_x} &                                                          &                                                          &                                                          &                                                          & \meter{} \\
\lstick{$q_{4}$} & \gate{R_x} &                                                          &                                                          &                                                          &                                                          & \meter{} \\
\lstick{$q_{5}$} & \gate{R_x} &                                                          &                                                          &                                                          &                                                          & \meter{} \\
\lstick{$q_{6}$} & \gate{R_x} &                                                          &                                                          &                                                          &                                                          & \meter{} \\
\lstick{$q_{7}$} & \gate{R_x} &                                                          &                                                          &                                                          &                                                          & \meter{} \\
\lstick{$q_{8}$} & \gate{R_x} &                                                          &                                                          &                                                          &                                                          & \meter{} \\
\lstick{$q_{9}$} & \gate{R_x} &                                                          &                                                          &                                                          &                                                          & \meter{}
\end{quantikz}
}%
};

\node[below=2mm of qhead, align=center] {\small Quantum Head};

\node[
    draw,
    fill=white,
    rectangle,
    inner sep=2mm,
    right=of qhead
] (chead) {%
\begin{tikzpicture}[
    scale=0.9,
    baseline=(current bounding box.center),
    layer1/.style={circle, draw, fill=red!40, minimum size=5mm, inner sep=0pt},
    layer2/.style={circle, draw, fill=yellow!30, minimum size=5mm, inner sep=0pt},
    layer3/.style={circle, draw, fill=green!30, minimum size=5mm, inner sep=0pt},
    conn/.style={thin}
]

\node[layer1] (i1) at (0,  1.6) {};
\node[layer1] (i2) at (0,  0.8) {};
\node[layer1] (i3) at (0,  0.0) {};
\node at (0,-0.6) {\Huge $\vdots$};
\node[layer1] (i4) at (0, -1.6) {};

\node[layer2] (h1) at (1.0,  1.2) {};
\node[layer2] (h2) at (1.0,  0.4) {};
\node at (1.0,-0.4) {\Huge $\vdots$};
\node[layer2] (h3) at (1.0, -1.2) {};

\node[layer3] (o1) at (2.0,  0.8) {};
\node at (2.0, 0) {\Huge $\vdots$};
\node[layer3] (o2) at (2.0, -0.8) {};

\foreach \a in {1,2,3,4}
    \foreach \b in {1,2,3}
        \draw[conn] (i\a) -- (h\b);

\foreach \a in {1,2,3}
    \foreach \b in {1,2}
        \draw[conn] (h\a) -- (o\b);

\end{tikzpicture}%
};

\node[below=2mm of chead, align=center] {\small Classical Head};

\node[block, fill=orange!30, right=of chead] (gfusion) {Gated\\Fusion};

\node[below=2mm of gfusion, align=center] {\small Mixing Parameter $\alpha$};

\node[right=of gfusion] (logits) {$\hat{\mathbf{y}}$};

\begin{scope}[on background layer]
    \draw[->, thick] ([xshift=7mm]img.east) -- (resnet.west);

    \coordinate (split) at ($(resnet.east)!0.4!(qhead.west)$);
    \draw[thick] (resnet.east) -- node[above] {$\mathbf{f}$} (split);

    \coordinate (top1) at ([yshift=5mm]qhead.north -| split);
    \coordinate (top2) at ([xshift=-5mm]chead.west |- top1);
    \coordinate (top3) at ([xshift=-5mm,yshift=8mm]chead.west);
    \coordinate (top4) at ([yshift=8mm]chead.west);

    \draw[->, thick]
        (split) -- (top1) -- (top2) -- (top3) -- (top4);

    \draw[->, thick] (split) -- (qhead.west);

    \draw[->, thick]
        (qhead.east)
        -- ([xshift=5mm]qhead.east)
        -- ([xshift=5mm, yshift=-8mm]qhead.east |- chead.south)
        -- ([xshift=-5mm, yshift=-7mm]gfusion.south -| gfusion.west)
        -- ([xshift=-5mm, yshift=-8mm]gfusion.west)
        -- ([yshift=-8mm]gfusion.west);

    \node[above, xshift=-2mm] at ($([xshift=-5mm, yshift=-8mm]gfusion.west)!0.5!([yshift=-8mm]gfusion.west)$) {$\mathcal{Q}(\mathbf{f})$};

    \draw[->, thick] ([yshift=8mm]chead.east) -- ([yshift=8mm]gfusion.west);

    \node[above, xshift=0.5mm] at ($([yshift=8mm]chead.east)!0.6!([yshift=8mm]gfusion.west)$) {$\mathcal{C}(\mathbf{f})$};

    \draw[->, thick] (gfusion.east) -- (logits.west);
\end{scope}

\end{tikzpicture}%
}%
}

\caption{\emph{QuantumBoostNet} overview: Feature vector $\mathbf{f}$ is routed in parallel to a quantum and a classical head, whose predictions are fused via $\alpha$; training proceeds in two phases, with the transition triggered by validation-accuracy plateau or exhaustion of the Phase~1 budget~$T_1$, whichever occurs first.}
\label{fig:ov}
\end{figure*}
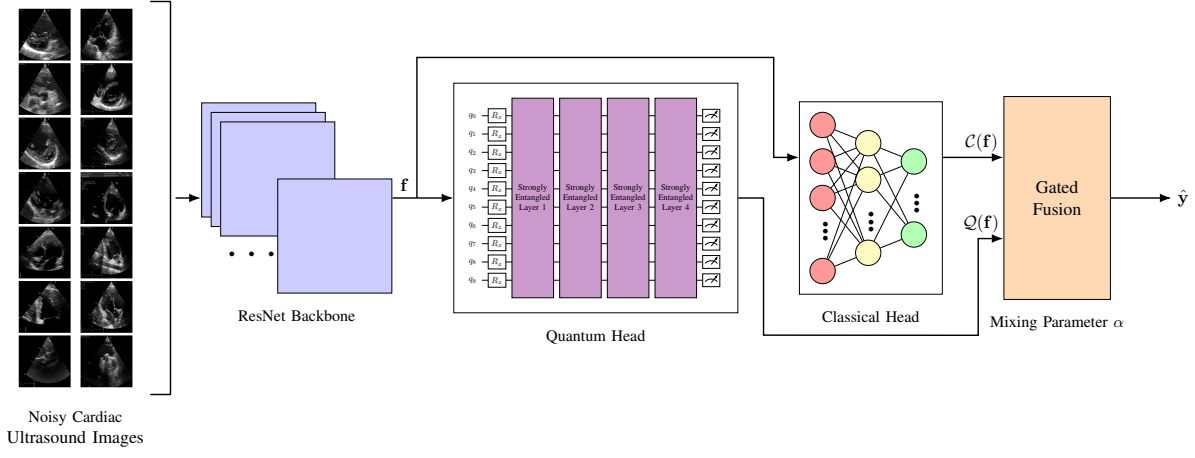

Addressing the limitations described above, \emph{QuantumBoostNet} has three variants (\emph{V1}--\emph{V3}), each using a two-phase training paradigm with quantum and classical prediction paths and dynamic phase switching. Depending on the variant and training phase, inference is either quantum-only or uses fused dual-path predictions. All variants share a common architecture but differ in managing training phases and protocols. Key differences include the order of training each prediction path, the timing of fusion activation, the parameterization and updating of the mixing coefficient, the role of the frozen path in the computation graph, and the conditioned phase switch. Figure~\ref{fig:ov} presents an overview of \emph{QuantumBoostNet}.

\emph{V1}--\emph{V3} have identical ResNet-18 backbones operating on single-channel (grayscale) inputs $\mathbf{X} \in \mathbb{R}^{1 \times H \times W}$, with $H$ and $W$ the input height and width. The initial feature map is
\begin{equation}\label{eq:qboost_stem}
  \mathbf{h}_0
  = \operatorname{MaxPool}_{3\times3,\,s{=}2}\!\Big(
      \operatorname{ReLU}\!\big(
        \operatorname{BN}\!\big(
          \operatorname{Conv}_{7\times7,\,s{=}2}(\mathbf{X})
        \big)
      \big)
    \Big),
\end{equation}
where $\mathbf{h}_0 \in \mathbb{R}^{64 \times \frac{H}{4} \times \frac{W}{4}}.$

The residual backbone has four stages, indexed $r$=$1,\dots,4$, each comprising two residual blocks; $\mathbf{h}_r$ is the output of residual stage $r$. A residual block with input $\mathbf{z}$ computes
\begin{equation}\label{eq:qboost_resblock_F}
\mathcal{F}(\mathbf{z})
=
\operatorname{BN}\!\Big(
\operatorname{Conv}_{3\times3,\,s{=}1}\!\big(
\operatorname{ReLU}\!\big(
\operatorname{BN}\!\big(
\operatorname{Conv}_{3\times3,\,s}(\mathbf{z})
\big)
\big)
\big)
\Big),
\end{equation}
\begin{equation}\label{eq:qboost_resblock}
\operatorname{ResBlock}(\mathbf{z})
=
\operatorname{ReLU}\!\big(\mathcal{F}(\mathbf{z}) + \mathcal{S}(\mathbf{z})\big),
\end{equation}
where $\mathcal{F}(\mathbf{z})$ is the residual branch transformation (i.e., two $3 \times 3$ convolutions with batch normalization, with a ReLU nonlinearity inserted in between). The first convolution uses stride $s$; the second always uses $1$. For blocks that preserve spatial resolution, $s=1$; for the first block of stages 2--4, $s=2$. The shortcut mapping $\mathcal{S}(\mathbf{z})$ is the identity when the input and output dimensions match. Otherwise, it is implemented as a $1 \times 1$ projection followed by batch normalization, using the same stride as the residual branch. Across the four stages, the channel dimensions progress as $64 \to 128 \to 256 \to 512$.

The final backbone representation is obtained by globally pooling the output of the fourth residual stage and flattening it into a shared feature vector
\begin{equation}\label{eq:qboost_feat}
  \mathbf{f}
  = \operatorname{Flatten}\!\big(
      \operatorname{AdaptiveAvgPool}_{1\times1}(\mathbf{h}_4)
    \big)
  \in \mathbb{R}^{512},
\end{equation}
where $\mathbf{h}_4$ is the output of the fourth residual stage. The adaptive average pooling operation aggregates each of the 512 feature channels into a single scalar, yielding a $1 \times 1$ spatial representation per channel. Flattening this tensor produces the shared feature vector $\mathbf{f} \in \mathbb{R}^{512}$, which is subsequently fed to both the quantum and classical branches.


The quantum path maps $\mathbf{f}$ through a classical dimensionality reduction, a variational quantum circuit, and a classical readout. The quantum branch first projects the shared backbone feature vector $\mathbf{f} \in \mathbb{R}^{512}$ into a circuit-input vector $\mathbf{u} \in \mathbb{R}^{N}$, where $N = 10$ is the number of qubits. The projection is followed by a $\tanh$ squashing and scaling by $\pi$,
\begin{equation}\label{eq:qboost_qin}
  \mathbf{u}
  = \pi \tanh\!\big(\mathbf{W}_{\mathrm{in}}\mathbf{f}
    + \mathbf{b}_{\mathrm{in}}\big)
  \in [-\pi,\pi]^N,
\end{equation}
where $\mathbf{W}_{\mathrm{in}} \in \mathbb{R}^{N \times 512}$ and $\mathbf{b}_{\mathrm{in}} \in \mathbb{R}^{N}$ are the learnable input projection parameters. Despite the information bottleneck, the classical ResNet backbone sufficiently condenses the relevant features before the quantum projection.

The variational quantum circuit acts on the initial state $\ket{0}^{\otimes N}$. The embedding layer is implemented with the PennyLane function \texttt{AngleEmbedding} without specifying the rotation argument, so the encoded features are $R_X$ rotations,
\begin{equation}\label{eq:qboost_angle_embed}
  \ket{\psi^{(0)}(\mathbf{u})}
  =
  \Bigg(
    \prod_{j=0}^{N-1} R_X(u_j)
  \Bigg)\ket{0}^{\otimes N},
\end{equation}
where $\mathbf{u} = [u_0,\dots,u_{N-1}]^\top \in \mathbb{R}^N$ is the vector of embedding angles produced by the classical input projection, $u_j$ is its $j$th component applied to qubit $j$, and $\ket{\psi^{(0)}(\mathbf{u})}$ is the state obtained after angle embedding from the initial state $\ket{0}^{\otimes N}$.
 
After embedding, the circuit applies $L$=4 variational layers of the PennyLane function \texttt{StronglyEntanglingLayers}. Let $\boldsymbol{\Omega} \in \mathbb{R}^{L \times N \times 3}$ denote the trainable template parameters, and let $(\phi_{\ell,j}, \theta_{\ell,j}, \omega_{\ell,j})$ be the three angles associated with qubit $j$ in layer $\ell$. Each layer first applies a single-qubit $\operatorname{Rot}(\phi_{\ell,j},\theta_{\ell,j},\omega_{\ell,j})$ gate to every wire, followed by a fixed entangling pattern,
\begin{equation}\label{eq:qboost_vqc_layer}
  \ket{\psi^{(\ell)}}
  =
  \mathcal{E}_{\ell}
  \Bigg(
    \prod_{j=0}^{N-1}
    \operatorname{Rot}(\phi_{\ell,j},\theta_{\ell,j},\omega_{\ell,j})
  \Bigg)
  \ket{\psi^{(\ell-1)}},
\end{equation}
where $\ell$ = 1,$\dots$, $L$, and
  $\operatorname{Rot}(\phi,\theta,\omega)$
  = $R_Z(\omega)\,R_Y(\theta)\,R_Z(\phi)$.
$\mathcal{E}_{\ell}$ denotes the entangling sublayer composed of non-parametric two-qubit gates. In PennyLane's default configuration for \texttt{StronglyEntanglingLayers}, these entanglers are CNOT gates applied according to the template-defined range pattern.

The quantum layer outputs one expectation value per qubit, obtained by measuring Pauli-$Z$ on every wire,
\begin{equation}\label{eq:qboost_qmeasure}
  m_j
  =
  \bra{\psi^{(L)}} Z_j \ket{\psi^{(L)}},
  \quad j=0,\dots,N-1,
\end{equation}
which yields the measurement vector $\mathbf{m} \in \mathbb{R}^{N}$, where $\mathbf{m} = [m_0,\dots,m_{N-1}]^\top$. 

\begin{figure*}[!t]
\centering

\resizebox{\textwidth}{!}{%
\begin{quantikz}[row sep={0.75cm,between origins}, column sep=0.1cm, thin lines]
\lstick{$q_{0}$} & \gate{R_x(x_0)} & \gate{\mathrm{R(\alpha^1_{1},\phantom{0} \beta^1_{1},\phantom{0} \gamma^1_{1})}}  & \ctrl{1} & \qw      & \qw      & \qw      & \qw      & \qw      & \qw      & \qw      & \qw      & \targ{}   & \gate{\mathrm{R(\alpha^2_{1},\phantom{0} \beta^2_{1},\phantom{0} \gamma^2_{1})}}  & \ctrl{2} & \qw      & \qw      & \qw      & \qw      & \qw      & \qw      & \qw      & \targ{}   & \qw       & \gate{\mathrm{R(\alpha^3_{1},\phantom{0} \beta^3_{1},\phantom{0} \gamma^3_{1})}}  & \ctrl{3} & \qw      & \qw      & \qw      & \qw      & \qw      & \qw      & \targ{}   & \qw       & \qw       & \gate{\mathrm{R(\alpha^4_{1},\phantom{0} \beta^4_{1},\phantom{0} \gamma^4_{1})}}  & \ctrl{4} & \qw      & \qw      & \qw      & \qw      & \qw      & \targ{}   & \qw       & \qw       & \qw       & \meter{} \\
\lstick{$q_{1}$} & \gate{R_x(x_1)} & \gate{\mathrm{R(\alpha^1_{2},\phantom{0} \beta^1_{2},\phantom{0} \gamma^1_{2})}}  & \targ{}  & \ctrl{1} & \qw      & \qw      & \qw      & \qw      & \qw      & \qw      & \qw      & \qw       & \gate{\mathrm{R(\alpha^2_{2},\phantom{0} \beta^2_{2},\phantom{0} \gamma^2_{2})}}  & \qw      & \ctrl{2} & \qw      & \qw      & \qw      & \qw      & \qw      & \qw      & \qw       & \targ{}   & \gate{\mathrm{R(\alpha^3_{2},\phantom{0} \beta^3_{2},\phantom{0} \gamma^3_{2})}}  & \qw      & \ctrl{3} & \qw      & \qw      & \qw      & \qw      & \qw      & \qw       & \targ{}   & \qw       & \gate{\mathrm{R(\alpha^4_{2},\phantom{0} \beta^4_{2},\phantom{0} \gamma^4_{2})}}  & \qw      & \ctrl{4} & \qw      & \qw      & \qw      & \qw      & \qw       & \targ{}   & \qw       & \qw       & \meter{} \\
\lstick{$q_{2}$} & \gate{R_x(x_2)} & \gate{\mathrm{R(\alpha^1_{3},\phantom{0} \beta^1_{3},\phantom{0} \gamma^1_{3})}}  & \qw      & \targ{}  & \ctrl{1} & \qw      & \qw      & \qw      & \qw      & \qw      & \qw      & \qw       & \gate{\mathrm{R(\alpha^2_{3},\phantom{0} \beta^2_{3},\phantom{0} \gamma^2_{3})}}  & \targ{}  & \qw      & \ctrl{2} & \qw      & \qw      & \qw      & \qw      & \qw      & \qw       & \qw       & \gate{\mathrm{R(\alpha^3_{3},\phantom{0} \beta^3_{3},\phantom{0} \gamma^3_{3})}}  & \qw      & \qw      & \ctrl{3} & \qw      & \qw      & \qw      & \qw      & \qw       & \qw       & \targ{}   & \gate{\mathrm{R(\alpha^4_{3},\phantom{0} \beta^4_{3},\phantom{0} \gamma^4_{3})}}  & \qw      & \qw      & \ctrl{4} & \qw      & \qw      & \qw      & \qw       & \qw       & \targ{}   & \qw       & \meter{} \\
\lstick{$q_{3}$} & \gate{R_x(x_3)} & \gate{\mathrm{R(\alpha^1_{4},\phantom{0} \beta^1_{4},\phantom{0} \gamma^1_{4})}}  & \qw      & \qw      & \targ{}  & \ctrl{1} & \qw      & \qw      & \qw      & \qw      & \qw      & \qw       & \gate{\mathrm{R(\alpha^2_{4},\phantom{0} \beta^2_{4},\phantom{0} \gamma^2_{4})}}  & \qw      & \targ{}  & \qw      & \ctrl{2} & \qw      & \qw      & \qw      & \qw      & \qw       & \qw       & \gate{\mathrm{R(\alpha^3_{4},\phantom{0} \beta^3_{4},\phantom{0} \gamma^3_{4})}}  & \targ{}  & \qw      & \qw      & \ctrl{3} & \qw      & \qw      & \qw      & \qw       & \qw       & \qw       & \gate{\mathrm{R(\alpha^4_{4},\phantom{0} \beta^4_{4},\phantom{0} \gamma^4_{4})}}  & \qw      & \qw      & \qw      & \ctrl{4} & \qw      & \qw      & \qw       & \qw       & \qw       & \targ{}   & \meter{} \\
\lstick{$q_{4}$} & \gate{R_x(x_4)} & \gate{\mathrm{R(\alpha^1_{5},\phantom{0} \beta^1_{5},\phantom{0} \gamma^1_{5})}}  & \qw      & \qw      & \qw      & \targ{}  & \ctrl{1} & \qw      & \qw      & \qw      & \qw      & \qw       & \gate{\mathrm{R(\alpha^2_{5},\phantom{0} \beta^2_{5},\phantom{0} \gamma^2_{5})}}  & \qw      & \qw      & \targ{}  & \qw      & \ctrl{2} & \qw      & \qw      & \qw      & \qw       & \qw       & \gate{\mathrm{R(\alpha^3_{5},\phantom{0} \beta^3_{5},\phantom{0} \gamma^3_{5})}}  & \qw      & \targ{}  & \qw      & \qw      & \ctrl{3} & \qw      & \qw      & \qw       & \qw       & \qw       & \gate{\mathrm{R(\alpha^4_{5},\phantom{0} \beta^4_{5},\phantom{0} \gamma^4_{5})}}  & \targ{}  & \qw      & \qw      & \qw      & \ctrl{4} & \qw      & \qw       & \qw       & \qw       & \qw       & \meter{} \\
\lstick{$q_{5}$} & \gate{R_x(x_5)} & \gate{\mathrm{R(\alpha^1_{6},\phantom{0} \beta^1_{6},\phantom{0} \gamma^1_{6})}}  & \qw      & \qw      & \qw      & \qw      & \targ{}  & \ctrl{1} & \qw      & \qw      & \qw      & \qw       & \gate{\mathrm{R(\alpha^2_{6},\phantom{0} \beta^2_{6},\phantom{0} \gamma^2_{6})}}  & \qw      & \qw      & \qw      & \targ{}  & \qw      & \ctrl{2} & \qw      & \qw      & \qw       & \qw       & \gate{\mathrm{R(\alpha^3_{6},\phantom{0} \beta^3_{6},\phantom{0} \gamma^3_{6})}}  & \qw      & \qw      & \targ{}  & \qw      & \qw      & \ctrl{3} & \qw      & \qw       & \qw       & \qw       & \gate{\mathrm{R(\alpha^4_{6},\phantom{0} \beta^4_{6},\phantom{0} \gamma^4_{6})}}  & \qw      & \targ{}  & \qw      & \qw      & \qw      & \ctrl{4} & \qw       & \qw       & \qw       & \qw       & \meter{} \\
\lstick{$q_{6}$} & \gate{R_x(x_6)} & \gate{\mathrm{R(\alpha^1_{7},\phantom{0} \beta^1_{7},\phantom{0} \gamma^1_{7})}}  & \qw      & \qw      & \qw      & \qw      & \qw      & \targ{}  & \ctrl{1} & \qw      & \qw      & \qw       & \gate{\mathrm{R(\alpha^2_{7},\phantom{0} \beta^2_{7},\phantom{0} \gamma^2_{7})}}  & \qw      & \qw      & \qw      & \qw      & \targ{}  & \qw      & \ctrl{2} & \qw      & \qw       & \qw       & \gate{\mathrm{R(\alpha^3_{7},\phantom{0} \beta^3_{7},\phantom{0} \gamma^3_{7})}}  & \qw      & \qw      & \qw      & \targ{}  & \qw      & \qw      & \ctrl{3} & \qw       & \qw       & \qw       & \gate{\mathrm{R(\alpha^4_{7},\phantom{0} \beta^4_{7},\phantom{0} \gamma^4_{7})}}  & \qw      & \qw      & \targ{}  & \qw      & \qw      & \qw      & \ctrl{-6} & \qw       & \qw       & \qw       & \meter{} \\
\lstick{$q_{7}$} & \gate{R_x(x_7)} & \gate{\mathrm{R(\alpha^1_{8},\phantom{0} \beta^1_{8},\phantom{0} \gamma^1_{8})}}  & \qw      & \qw      & \qw      & \qw      & \qw      & \qw      & \targ{}  & \ctrl{1} & \qw      & \qw       & \gate{\mathrm{R(\alpha^2_{8},\phantom{0} \beta^2_{8},\phantom{0} \gamma^2_{8})}}  & \qw      & \qw      & \qw      & \qw      & \qw      & \targ{}  & \qw      & \ctrl{2} & \qw       & \qw       & \gate{\mathrm{R(\alpha^3_{8},\phantom{0} \beta^3_{8},\phantom{0} \gamma^3_{8})}}  & \qw      & \qw      & \qw      & \qw      & \targ{}  & \qw      & \qw      & \ctrl{-7} & \qw       & \qw       & \gate{\mathrm{R(\alpha^4_{8},\phantom{0} \beta^4_{8},\phantom{0} \gamma^4_{8})}}  & \qw      & \qw      & \qw      & \targ{}  & \qw      & \qw      & \qw       & \ctrl{-6} & \qw       & \qw       & \meter{} \\
\lstick{$q_{8}$} & \gate{R_x(x_8)} & \gate{\mathrm{R(\alpha^1_{9},\phantom{0} \beta^1_{9},\phantom{0} \gamma^1_{9})}}  & \qw      & \qw      & \qw      & \qw      & \qw      & \qw      & \qw      & \targ{}  & \ctrl{1} & \qw       & \gate{\mathrm{R(\alpha^2_{9},\phantom{0} \beta^2_{9},\phantom{0} \gamma^2_{9})}}  & \qw      & \qw      & \qw      & \qw      & \qw      & \qw      & \targ{}  & \qw      & \ctrl{-8} & \qw       & \gate{\mathrm{R(\alpha^3_{9},\phantom{0} \beta^3_{9},\phantom{0} \gamma^3_{9})}}  & \qw      & \qw      & \qw      & \qw      & \qw      & \targ{}  & \qw      & \qw       & \ctrl{-7} & \qw       & \gate{\mathrm{R(\alpha^4_{9},\phantom{0} \beta^4_{9},\phantom{0} \gamma^4_{9})}}  & \qw      & \qw      & \qw      & \qw      & \targ{}  & \qw      & \qw       & \qw       & \ctrl{-6} & \qw       & \meter{} \\
\lstick{$q_{9}$} & \gate{R_x(x_9)} & \gate{\mathrm{R(\alpha^1_{10},           \beta^1_{10},           \gamma^1_{10})}} & \qw      & \qw      & \qw      & \qw      & \qw      & \qw      & \qw      & \qw      & \targ{}  & \ctrl{-9} & \gate{\mathrm{R(\alpha^2_{10},           \beta^2_{10},           \gamma^2_{10})}} & \qw      & \qw      & \qw      & \qw      & \qw      & \qw      & \qw      & \targ{}  & \qw       & \ctrl{-8} & \gate{\mathrm{R(\alpha^3_{10},           \beta^3_{10},           \gamma^3_{10})}} & \qw      & \qw      & \qw      & \qw      & \qw      & \qw      & \targ{}  & \qw       & \qw       & \ctrl{-7} & \gate{\mathrm{R(\alpha^4_{10},           \beta^4_{10},           \gamma^4_{10})}} & \qw      & \qw      & \qw      & \qw      & \qw      & \targ{}  & \qw       & \qw       & \qw       & \ctrl{-6} & \meter{}
\end{quantikz}
}%

\caption{
Variational quantum circuit shared by the \texttt{QuantumBoostNet} models (V1, V2, and V3), with $n = 10$ qubits and $L = 4$ layers.
Classical features $\mathbf{x}$ are encoded using the \texttt{AngleEmbedding} function, using $R_x$ gates, applying $R_x(x_i)$ to each qubit~$i$ (where $i = 0, 1, \dots, 9$).
Each variational layer applies a $R(\alpha, \beta, \gamma) = R_z(\gamma)\, R_y(\beta)\, R_z(\alpha)$ gate (three trainable parameters per qubit per layer).
It is then followed by a ring of CNOT gates from the \texttt{StronglyEntanglingLayers} function.
In layer~$\ell$ ($\ell = 0, \dots, 3$), each qubit~$i$ controls qubit $(i + r_\ell) \bmod n$ with range $r_\ell = (\ell \bmod (n - 1)) + 1$, yielding $r \in \{1,2,3,4\}$.
Each qubit is measured in the Pauli-$Z$ basis, yielding the expectation values.
}

\label{fig:vqc}
\end{figure*}
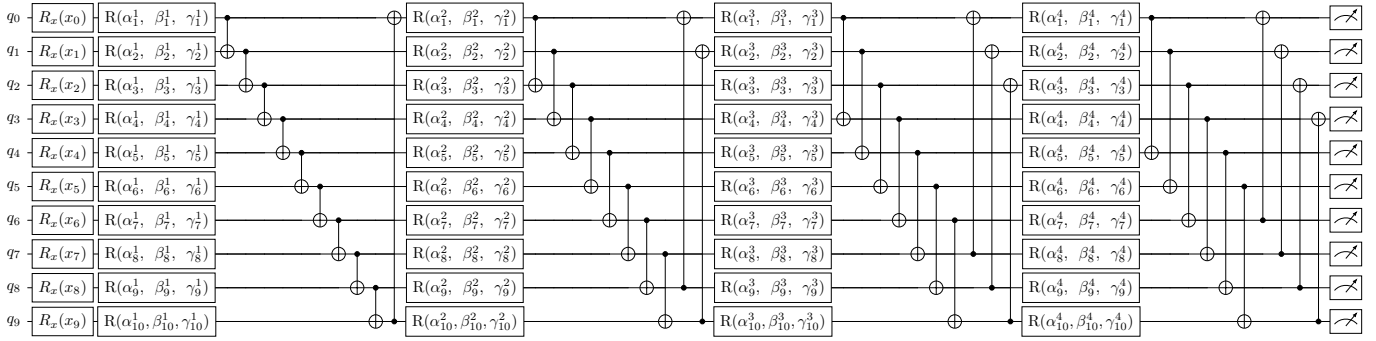

Finally, a classical linear readout maps these expectation values to class logits
\begin{equation}\label{eq:qboost_qhead}
  \mathcal{Q}(\mathbf{f})
  =
  \mathbf{W}_{\mathrm{out}}\mathbf{m}
  + \mathbf{b}_{\mathrm{out}}
  \in \mathbb{R}^{C},
\end{equation}
where $\mathbf{W}_{\mathrm{out}} \in \mathbb{R}^{C \times N}$ and $\mathbf{b}_{\mathrm{out}} \in \mathbb{R}^{C}$ are learnable output parameters, and $C$ is the number of classes. Fig.~\ref{fig:vqc} shows the variational quantum circuit architecture for \emph{QuantumBoostNet}. 

The shared feature vector $\mathbf{f}$ is processed by a classical classification head, separate from the quantum branch. The classical branch $\mathcal{C}(\mathbf{f})$ is a two-layer MLP with an intermediate 256-dimensional hidden representation, followed by batch normalization, ReLU, and dropout with probability $0.3$,
\begin{equation}\label{eq:classical_head}
    \mathbf{W}_2 \Big(
      \operatorname{Dropout}_{p=0.3}\!\big(
        \operatorname{ReLU}\!\big(
          \operatorname{BN}\!\big(
            \mathbf{W}_1\,\mathbf{f} + \mathbf{b}_1
          \big)
        \big)
      \big)
    \Big) + \mathbf{b}_2.
\end{equation}
$\mathcal{C}(\mathbf{f}) \in \mathbb{R}^{C}$,
$\mathbf{W}_1 \in \mathbb{R}^{256 \times 512}$ and $\mathbf{b}_1 \in \mathbb{R}^{256}$ are the weight matrix and bias vector of the first linear layer;
$\mathbf{W}_2 \in \mathbb{R}^{C \times 256}$, $\mathbf{b}_2 \in \mathbb{R}^{C}$ are the weight and bias of the output layer.


The predictions of the quantum and classical branches are combined differently across the three \emph{QuantumBoostNet} variants. Let $\mathcal{Q}(\mathbf{f}) \in \mathbb{R}^{C}$ denote the quantum logits and $\mathcal{C}(\mathbf{f}) \in \mathbb{R}^{C}$ the classical logits. In \emph{V1} and \emph{V3} of \emph{QuantumBoostNet}, as well as in the boost phase of \emph{V2}, the final logits are obtained by decision-level fusion
\begin{equation}\label{eq:gated_fusion}
  \hat{\mathbf{y}}
  = \alpha\,\mathcal{Q}(\mathbf{f})
    + (1-\alpha)\,\mathcal{C}(\mathbf{f}),
  \quad \alpha \in (0,1),
\end{equation}
where $\hat{\mathbf{y}} \in \mathbb{R}^{C}$ denotes the fused output logits.

Parameter $\alpha$ controls the relative contribution of the quantum and classical branches. In \emph{V1} and \emph{V3}, $\alpha$ is induced by a learnable scalar gate through a sigmoid transformation, and is later steered during the boost phase according to the variant-specific schedule. In \emph{V2}, fusion is inactive during the initial quantum-only phase,
  $\hat{\mathbf{y}}$ = $\mathcal{Q}(\mathbf{f})$,
and is introduced only in the second phase through the mixing coefficient $\alpha$.


For our novel training protocol, we define the nominal total training budget as $T = T_1 + T_2$ epochs, split into Phase~1 and Phase~2. Here, $t$ denotes the current training epoch. Let $t^{*}$ denote the actual transition epoch, triggered by whichever occurs first: plateau detection (i.e., the validation accuracy improvement falls below $\epsilon = 0.002$, where $\epsilon$ is the minimum accuracy improvement required to avoid declaring a plateau, for two consecutive epochs after a 5-epoch warm-up), or exhaustion of the nominal Phase~1 budget~$T_1$. Specifically,
\begin{equation}\label{eq:plateau}
  \Delta_t
  = \mathrm{Acc}_{\mathrm{val}}^{(t)}
    - \mathrm{Acc}_{\mathrm{val}}^{(t-1)},
\end{equation}
and plateau is declared when $\Delta_t$ $<$ $\epsilon$ for two consecutive epochs with $t>5$.

At $t = t^{*}$, the optimizer is reinitialized as Adam with learning rate $\eta_0 = 10^{-3}$ over all currently trainable parameters, and a fresh cosine-annealing schedule is started for Phase~2 over the nominal Phase~2 duration budget $T_2$.

\subsection{Variants description and implementation}

We denote the parameter sets by
$\Theta_{\mathrm{backbone}}$ for the shared ResNet-18-style feature extractor,
$\Theta_Q$ for the quantum path,
$\Theta_{\mathcal{C}}$ for the classical head,
and $\{g\}$ for the scalar gate parameter when applicable.
Precisely,
$\Theta_Q$ comprises the input projection to the quantum register, the
variational quantum-circuit parameters, and the quantum readout layer,
while $\Theta_{\mathcal{C}}$ contains the two linear layers of the
classical head together with the batch-normalization parameters.


Although all three \emph{QuantumBoostNet} variants share the same backbone, quantum path, and classical head, they differ in 
four implementation-level design choices:
(i) which branch is optimized first,
(ii) whether fusion is active from the beginning or only after the phase transition,
(iii) how the mixing coefficient $\alpha$ is represented and updated, and
(iv) whether the frozen branch remains inside the computation graph or is explicitly detached.


{\bf QuantumBoostNet V1} starts from a quantum-dominant regime. The mixing coefficient is $\alpha = \sigma(g)$, where $g \in \mathbb{R}$ is a learnable scalar and $\sigma$ the sigmoid function. The initial gate value is $g_0$ = 3.0, which yields $\alpha_0$ = $\sigma(3.0) \approx 0.95$, so the fused prediction is initially dominated by the quantum branch.

The first phase uses the quantum head, with a frozen classical head; the shared backbone, the quantum path, and the gate parameter remain trainable. The active parameter set is $\Theta_{\mathrm{backbone}} \cup \Theta_Q \cup \{g\}$. Although the final logits are already formed through fusion, the large initial value of $\alpha$ strongly biases the model toward the quantum prediction. During this phase, the gate is learned jointly with the backbone and quantum parameters through backpropagation. 
        
The second phase deals with the classical head. Once the transition criterion is met, the quantum-specific path is frozen, and the classical head is activated. The active parameter set becomes $\Theta_{\mathrm{backbone}} \cup \Theta_{\mathcal{C}}$. At this point, the role of the classical head is to improve upon the representation learned during the quantum-first stage. The fusion coefficient is no longer free, but steered through a linear annealing rule,
        \begin{equation}\label{eq:v1_anneal}
          \alpha^{(t)}
          = \alpha^{(t^{*})}
            + \frac{t - t^{*}}{T - t^{*}}
              \bigl(0.2 - \alpha^{(t^{*})}\bigr),
          \qquad t > t^{*},
        \end{equation}
so that the contribution of the classical branch gradually increases throughout the second phase. Here, $\alpha^{(t^{*})}$ denotes the value of the mixing coefficient at the moment the Phase 1 to Phase 2 transition is triggered. The corresponding gate value is injected through the inverse-logit transform,
        \begin{equation}\label{eq:v1_logit}
          g^{(t)} = \log\!\frac{\alpha^{(t)}}{1 - \alpha^{(t)}}.
        \end{equation}

Accordingly, \emph{V}1 employs a soft quantum-first curriculum. Training first prioritizes the quantum pathway, allowing the backbone to adapt to quantum-driven supervision. When progress plateaus, the classical head is added as a secondary corrective branch. The fusion schedule then gradually shifts decision-making from quantum logits to a more classically informed approach.


{\bf QuantumBoostNet V2} clearly separates stages. Unlike \emph{V}1, the mixing coefficient is stored as a scalar and updated only by the training schedule, not by a learnable gate. The initial phase of \emph{V}2 acts as a purely quantum classifier, with no classical contribution during the forward pass.

Phase~1 is purely quantum, so the model uses only the quantum branch,
         $\hat{\mathbf{y}}^{(t)}$ = $\mathcal{Q}(\mathbf{f})$,
          $t \leq t^{*}$.
The classical head is neither evaluated nor updated, and the active parameter set is $\Theta_{\mathrm{backbone}} \cup \Theta_Q$. This phase is intended to train the shared feature extractor under purely quantum supervision. 

Phase~2 handles the classical part. After the switch, the quantum path is frozen, while the classical head is unfrozen. The active parameter set becomes $\Theta_{\mathrm{backbone}} \cup \Theta_{\mathcal{C}}$. The forward pass is then given by
        \begin{equation}\label{eq:v2_forward}
          \hat{\mathbf{y}}^{(t)}
          = \alpha^{(t)} \cdot
            \mathcal{Q}(\mathbf{f})
            + \bigl(1 - \alpha^{(t)}\bigr) \cdot \mathcal{C}(\mathbf{f}),
          \qquad t > t^{*},
        \end{equation}        
where $\mathcal{Q}(\mathbf{f})$ behaves like a fixed reference signal, not like a trainable path that still influences optimization through gradients, and the mixing coefficient follows the fixed linear schedule
        \begin{equation}\label{eq:v2_anneal}
          \alpha^{(t)}
          = 0.8 + \frac{t - t^{*}}{T - t^{*}}
            \bigl(0.2 - 0.8\bigr),
          \qquad t > t^{*}.
        \end{equation}
Thus, Phase~2 begins from a still quantum-favouring mixture, but progressively shifts weight toward the classical head.
Because the quantum branch is detached from the computation graph during Phase 2, gradients propagate only through the classical branch and the shared backbone. This prevents direct co-adaptation of the frozen quantum parameters.


{\bf QuantumBoostNet V3} mirrors the philosophy of \emph{V}1, but reverses the order of the two branches. As in \emph{V}1, the mixing coefficient is parameterized as $\alpha = \sigma(g)$ with a learnable scalar gate. The gate is initialized as $g_0$ = -3.0, yielding $\alpha_0$ = $\sigma(-3.0) \approx 0.05$. Thus, the fused prediction at the beginning of training is classical-dominant. During the first phase, the quantum path is frozen, while the shared backbone, the classical head, and the gate remain trainable. The active parameter set is $\Theta_{\mathrm{backbone}} \cup \Theta_{\mathcal{C}} \cup \{g\}$. The quantum branch is frozen and its contribution is mitigated by the small $\alpha$, so the model behaves as a classical classifier during this phase. The gate is still learnable, so the precise fusion ratio may adjust, but the forward pass remains dominated by the classical logits.

For the second phase, after the transition, the classical head is frozen and the quantum path is unfrozen. The active parameter set becomes $\Theta_{\mathrm{backbone}} \cup \Theta_Q$. The fusion coefficient is then annealed toward greater quantum emphasis,
        \begin{equation}\label{eq:v3_anneal}
          \alpha^{(t)}
          = \alpha^{(t^{*})}
            + \frac{t - t^{*}}{T - t^{*}}
              \bigl(0.8 - \alpha^{(t^{*})}\bigr),
          \qquad t > t^{*},
        \end{equation}
the gate value being set through the inverse-logit mapping.

\emph{V}3 implements a classical-first curriculum. The backbone is first shaped by the more easily optimized classical branch. Then, the quantum path is activated as a secondary enhancement. 


Taken together, \emph{V}1--\emph{V}3 form a family of staged hybrid models that share structural components but use distinct training paradigms. \emph{V}1 adopts a soft quantum-first strategy with learnable gating. \emph{V}2 implements a strict transition from quantum-only to quantum-plus-classical phases. \emph{V}3 inverts \emph{V}1's sequence. All variants were implemented within a unified codebase, sharing the backbone, quantum layer, and classical head definitions. They differ only in branch-freezing logic, gate handling, and phase-transition schedules.
The schedule constants were fixed in preliminary experiments. $g_0=\pm3.0$ gives near-exclusive Phase~1 dominance while keeping the sigmoid gate out of its saturated region, the annealing targets $0.2$ and $0.8$ leave the frozen branch a fixed $20\%$ contribution, $\epsilon=0.002$ matches run-to-run noise, and the five-epoch warm-up prevents premature switching.


\section{Experimental Results}

\subsection{Experimental environment and tested models}


Python version 3.12.12 was the main programming language and managed using Conda. The experiments were developed and carried out in Jupyter Notebook using ipykernel. The key libraries used were PennyLane 0.44.0, scikit-learn 1.8.0, torch 2.9.1, and torchvision 0.24.1. Experiments involving the View Classification dataset ran on a personal computer with Intel(R) Core(TM) i9-14900K, NVIDIA GeForce RTX 4070 Ti SUPER GPU, and 64 GB of RAM. Experiments on the FashionMNIST~\cite{xiao2017fashion} and MNIST~\cite{lecun1998gradient} were carried out on a different server with an Intel(R) Xeon(R) Gold 6240 CPU at 2.60 GHz, an NVIDIA Tesla T4 GPU, and 64 GB of RAM.


\subsubsection{Classical models}

Six classical architectures of increasing complexity are benchmarked for their ability to perform view identification on still images: 
~\emph{ViewCNN}, a lightweight three-block convolutional neural network (CNN);
~\emph{ViewOptimizedCNN}, a four-block convolutional neural network incorporating batch normalization and dropout;
~\emph{ViewResNet}~\cite{he2016resnet}, a ResNet-18 backbone with a two-layer fully connected head;
~\emph{ViewOptimizedResNet}~\cite{he2016resnet}, a custom residual backbone with four stages of 64--512 channels;
~\emph{MadaniCNN}, our implementation of the CNN architecture specialized for view identification~\cite{madani2018}; and
~\emph{LiViewNet}, our implementation of the view classification branch of the multi-task model of Li et al.~\cite{li2024multitask} (the quality assessment branch and its associated modules are omitted, as our dataset contains no quality labels).

These models establish a comprehensive baseline for comparison with hybrid classical-quantum models.


\subsubsection{Hybrid classical-quantum models}

Six hybrid classical-quantum model types were compared, each representing a distinct level of design sophistication. The first four models are direct hybridization approaches that adapt previously developed classical models. The same quantum circuit architecture is integrated into each of the following:
\emph{ViewQNN}, a three-block CNN;
\emph{ViewOptimizedQNN}, a deeper four-block CNN with batch normalization and dropout;
\emph{ViewQResNet}, which replaces the classical head of a ResNet-18 backbone with the quantum circuit; and
\emph{ViewOptimizedQResNet}, a custom residual backbone that contains a BatchNorm sandwich bridge to stabilize angle encoding prior to the variational circuit.


The remaining two architecture-level hybrid models incorporate more advanced structural concepts identified in the literature:
\emph{DualPathQuantumNet} maintains parallel classical and quantum feature-extraction paths, with their outputs concatenated prior to a shared classifier. This model incorporates focal loss~\cite{lin2017focal} and cosine-annealing scheduling. This model employs a quantum circuit architecture that is distinct from the previous four models.
\emph{PHN}, our implementation of the Parallel Hybrid Network proposed by Kordzanganeh et al.~\cite{kordzanganeh2023phn}, adapted for image input through the addition of a lightweight CNN feature extractor. The resulting feature vector is processed in parallel by a Variational Quantum Circuit (VQC) branch and a Multi-Layer Perceptron (MLP) branch, with their outputs fused using trainable per-output scalar weights. PHN serves as a conceptual precursor to the gated fusion mechanism in the \emph{QuantumBoostNet} architecture, but it trains both paths simultaneously with independent scalars.

\subsection{Benchmarks and datasets}


An extended version of the echocardiographic dataset originally presented by Wegner et al.~\cite{wegner2022accuracy} was utilized to investigate view classification in cases of altered cardiac anatomy. The dataset was collected at the Department of Cardiology~III, Adult Congenital and Valvular Heart Disease, University Hospital Muenster, Germany, with local ethics approval (Ärztekammer Westfalen-Lippe, no.~2020-751-f-S). The dataset by Wegner et al.~\cite{wegner2022accuracy} was selected due to its demonstrated challenges for state-of-the-art methods.

The dataset comprises two-dimensional TTE studies from 262 C/SHD patients (mean age $49\pm17$ years, 60\% male) spanning a wide range of congenital and structural pathologies, and $62$ structurally normal controls. Studies were acquired by multiple echocardiographers on GE (Vivid~7/E9/E95) and Philips (EPIQ~7C/7G, iE33) systems in inpatient and outpatient settings, introducing realistic variability in gain, depth, sector width, frame rate, and image quality.

Wegner et al. started from the $23$-view taxonomy used by Zhang et al.~\cite{zhang2018fully} but pruned it to $17$ view classes because several rarer views had too few examples in the C/SHD cohort to support robust learning. The retained $17$ classes cover the main acquisition windows of a comprehensive TTE examination as recommended by the American Society of Echocardiography \cite{mitchell2019guidelines} and are listed in Table~\ref{tab:views} (abbreviations: PLAX = parasternal long axis, PSAX = parasternal short axis, A2C = apical two chamber, A3C = apical three chamber, A4C = apical four chamber, MV = mitral valve, AV = aortic valve, RV = right ventricle). These classes represent a subset of standard TTE acquisition windows, broadly aligned with comprehensive adult TTE practice recommendations.

\begin{table}[t]
\centering
\caption{The seventeen transthoracic echocardiography view classes retained in the dataset of Wegner et al.~\cite{wegner2022accuracy}.}
\label{tab:views}
\footnotesize
\begin{tabular}{@{}cl@{\hspace{2em}}cl@{}}
\toprule
\# & View class                 & \# & View class \\
\midrule
1 & PLAX left ventricle         & 10 & A4C zoomed left ventricle \\
2 & PLAX zoomed MV              & 11 & Apical 5 chamber \\
3 & PLAX RV inflow              & 12 & A2C \\
4 & PSAX focus on AV            & 13 & A2C zoomed left ventricle \\
5 & PSAX papillary muscles      & 14 & A3C \\
6 & PSAX apex                   & 15 & A3C zoomed left ventricle \\
7 & PSAX zoomed AV              & 16 & Subcostal 4 chamber \\
8 & PSAX MV                     & 17 & Suprasternal aortic arch \\
9 & A4C                         &    & \\
\bottomrule
\end{tabular}
\end{table}

The dataset was refined to 14 view classes, each with an internal numeric label, by merging views that medical doctors cannot reliably distinguish on individual grayscale frames with their closest anatomical counterparts. Since our pipeline classifies single static frames, keeping such classes would introduce label noise instead of clinically meaningful distinctions. The mapping between these labels and view names is available (with the entire refined dataset) upon request. The imaging loops were decomposed into 210,345 individual frames. DICOM studies were anonymized, exported, and converted into individual grayscale PNG frames for automated analysis.
Frames are ensured to appear only in the training, validation or test partition, never two or more partitions; we also ensured a patient-level split.
This resulted in 175,146 frames for training and validation, and 35,199 for testing. All frames were resized to 224×224 pixels and reduced to 256 grayscale before training. 


We also evaluated whether our models surpass the implemented baselines in standard image classification benchmarks.
Therefore, we evaluated our models using the FashionMNIST~\cite{xiao2017fashion} and MNIST~\cite{lecun1998gradient} datasets. 

\subsection{Experimental setups}


Four distinct settings (S1--S4) were used to comprehensively characterize and evaluate model behavior. With the exception of S2, each model from every setting was evaluated with 10-fold cross-validation, with 15 epochs for each training.
All models, code, settings, and experimental results are available at \url{https://github.com/MihaiUM2001/CardioQML}.


{\bf Setting S1} uses the original View Classification dataset under clean conditions. The average test accuracy was recorded for each model. Then, the models were retrained to measure total training time and epoch duration.


{\bf Setting S2} uses the View Classification dataset to assess the \emph{QuantumBoostNet} models in a Noisy Intermediate-Scale Quantum (NISQ) environment by introducing gate-level noise. For training, noise was injected after each eligible gate with a probability of $p_{1q}$=0.005 for 1-qubit gates and a probability of $p_{2q}$=0.02 for 2-qubit gates. Two types of noises were considered using PennyLane, namely \texttt{Depolarizing} and \texttt{AmplitudeDamping}.
The models are trained for 15 epochs. Due to computational limitations and to maintain computational feasibility, the classical-quantum hybrid models were limited to 4 qubits and 2 variational layers, while respecting Equations \ref{eq:qboost_angle_embed}--\ref{eq:qboost_qmeasure}. The noisy experiments were conducted once per model and per noise channel, with both test accuracy and training time recorded.
Because each noisy configuration was trained once, these results should be read as indicative single-run measurements.
We also conducted inference and evaluation of a previously trained \emph{QuantumBoostNet~V1} from S1 using real NISQ IBM quantum hardware.
The IBM quantum hardware operates as a queue-based system, which makes submitting the entire test set simultaneously impractical. Therefore, we had to create a subset of our test set, selecting randomly only 1\% of images from each class (totalling 351 images), and using 73 minutes and 36 seconds of allocated time on the real hardware.


{\bf Settings S3 and S4} use the FashionMNIST~\cite{xiao2017fashion} and the MNIST~\cite{lecun1998gradient} datasets, respectively, to train the models. Apart from adapting the models to fit these datasets better, the training method is identical to S1.

\subsection{Performance comparison}


\subsubsection{Results for setting S1}

Table~\ref{tab:vc_clean} reports the mean test accuracy and standard deviation over 10-fold cross-validation for all 15 architectures on the cardiac ultrasound view classification dataset for S1. Average per-epoch training times, measured in a separate single-run experiment, are also provided.

\begin{table}[!t]
\centering
\caption{Average model accuracy (10-fold CV), average per-epoch training time, and average macros on the View Classification dataset for S1. Best scores, excluding training time, in bold.}
\label{tab:vc_clean}
\setlength{\tabcolsep}{3pt}
\renewcommand{\arraystretch}{1.05}
\resizebox{\columnwidth}{!}{%
\begin{tabular}{lccccccc}
\toprule
\textbf{Model} & \textbf{Accuracy (\%)} & \textbf{Time (s)} & \textbf{Precision (\%)} & \textbf{\shortstack{Weighted \\ Precision (\%)}} & \textbf{Recall (\%)} & \textbf{F1 (\%)} & \textbf{\shortstack{Weighted \\ F1 (\%)}} \\
\midrule
\emph{ViewCNN} & 53.55 $\pm$ 1.61 & 93.0 & 45.4 & 52.8 & 41.4 & 41.8 & 51.7 \\
\emph{ViewOptimizedCNN} & 63.44 $\pm$ 1.39 & 125.5 & 56.8 & 64.6 & 52.7 & 52.3 & 62.1 \\
\emph{ViewResNet} & 75.95 $\pm$ 1.18 & 124.8 & 72.9 & 75.7 & 68.8 & 69.5 & 75.5 \\
\emph{ViewOptimizedResNet} & 75.54 $\pm$ 1.23 & 124.2 & 72.7 & 75.4 & 69.2 & 69.9 & 75.0 \\
\emph{MadaniCNN} & 68.76 $\pm$ 1.35 & 253.4 & 61.6 & 68.1 & 58.7 & 59.3 & 68.2 \\
\emph{LiViewNet} & 75.52 $\pm$ 1.36 & 635.9 & 72.2 & 75.6 & 69.4 & 69.7 & 75.0 \\
\midrule
\emph{ViewQNN} & 51.28 $\pm$ 3.25 & 410.3 & 43.8 & 51.3 & 39.3 & 39.7 & 50.0 \\
\emph{ViewOptimizedQNN} & 52.78 $\pm$ 3.88 & 500.2 & 49.3 & 57.8 & 44.9 & 44.5 & 52.6 \\
\emph{ViewQResNet} & 73.23 $\pm$ 1.75 & 492.9 & 70.3 & 73.3 & 65.8 & 66.3 & 72.3 \\
\emph{ViewOptimizedQResNet} & 72.94 $\pm$ 1.83 & 490.9 & 68.0 & 73.0 & 64.4 & 64.8 & 72.6 \\
\emph{DualPathQuantumNet} & 74.21 $\pm$ 0.57 & 492.9 & \textbf{78.3} & 75.5 & 58.9 & 59.8 & 71.1 \\
\emph{PHN} & 54.84 $\pm$ 1.52 & 515.8 & 48.0 & 54.4 & 43.9 & 44.6 & 53.7 \\
\midrule
\emph{QuantumBoostNet V1} & \textbf{77.19 $\pm$ 0.89} & 426.9 & 75.1 & \textbf{76.8} & \textbf{70.8} & \textbf{71.8} & \textbf{76.7} \\
\emph{QuantumBoostNet V2} & 74.93 $\pm$ 2.02 & 429.3 & 71.4 & 74.6 & 67.6 & 68.3 & 74.3 \\
\emph{QuantumBoostNet V3} & 76.63 $\pm$ 0.67 & 410.9 & 74.6 & 76.3 & 69.1 & 70.2 & 75.9 \\
\bottomrule
\end{tabular}%
}
\end{table}

Among the classical baselines in S1, \emph{ViewResNet} achieves the highest accuracy at 75.95\%. An important thing to note is that a simple quantum-head substitution does not uniformly improve performance. \emph{ViewQNN} (51.28\%) underperforms even the weakest classical baseline, indicating that shallow CNN backbones do not produce features amenable to low-qubit variational classification. By contrast, ResNet-backed hybrid models, \emph{ViewQResNet}, \emph{ViewOptimizedQResNet}, and \emph{DualPathQuantumNet}, approach classical performance, suggesting that a sufficiently expressive backbone is a prerequisite for effective hybrid classification.

\begin{figure}[!t]
\centering
\includegraphics[scale=0.30]{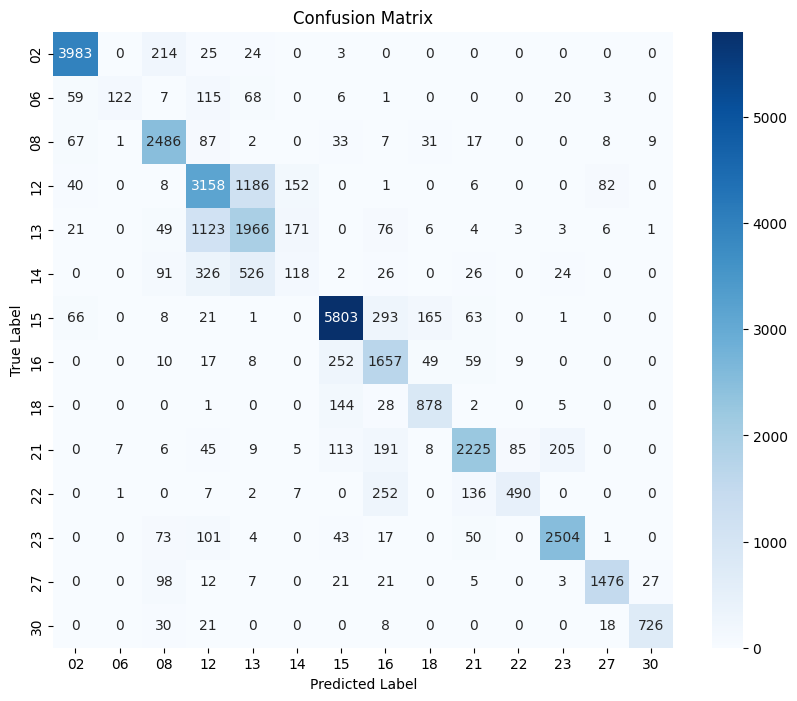}
\caption{The confusion matrix of \emph{QuantumBoostNet~V1's} sixth fold.}
\label{fig:vc_confusionmatrix}
\end{figure}

\emph{QuantumBoostNet~V1} achieves the best overall accuracy of 77.19 $\pm$ 0.89\%, a +1.24 percentage point (pp) absolute improvement over the best classical model (\emph{ViewResNet}). \emph{QuantumBoostNet~V3} follows at 76.63 $\pm$ 0.67\% with the lowest standard deviation of all \emph{QuantumBoostNet} models, indicating particularly stable fold-to-fold performance. \emph{QuantumBoostNet~V2} reaches 74.93\% $\pm$ 2.02\%, ranking above all non-\emph{QuantumBoostNet} hybrid models. Fig.~\ref{fig:vc_confusionmatrix} showcases the confusion matrix for the \emph{QuantumBoostNet~V1} with the highest test accuracy, which is the sixth fold.
All other confusion matrices are on \href{https://github.com/MihaiUM2001/CardioQML/blob/main/ViewClassification/VC_Default/ViewClassification.ipynb}{GitHub}.
Classical models require 93--635.9\,s per epoch, while hybrid models range from 410.3 to 515.8\,s per epoch. The \emph{QuantumBoostNet} variants train at 410.9--429.3\,s per epoch, placing them at the lower end of the hybrid range while delivering the highest accuracy and representing a favorable accuracy--cost trade-off within the hybrid family.

\begin{figure}[!t]
\centering
\includegraphics[scale=0.25]{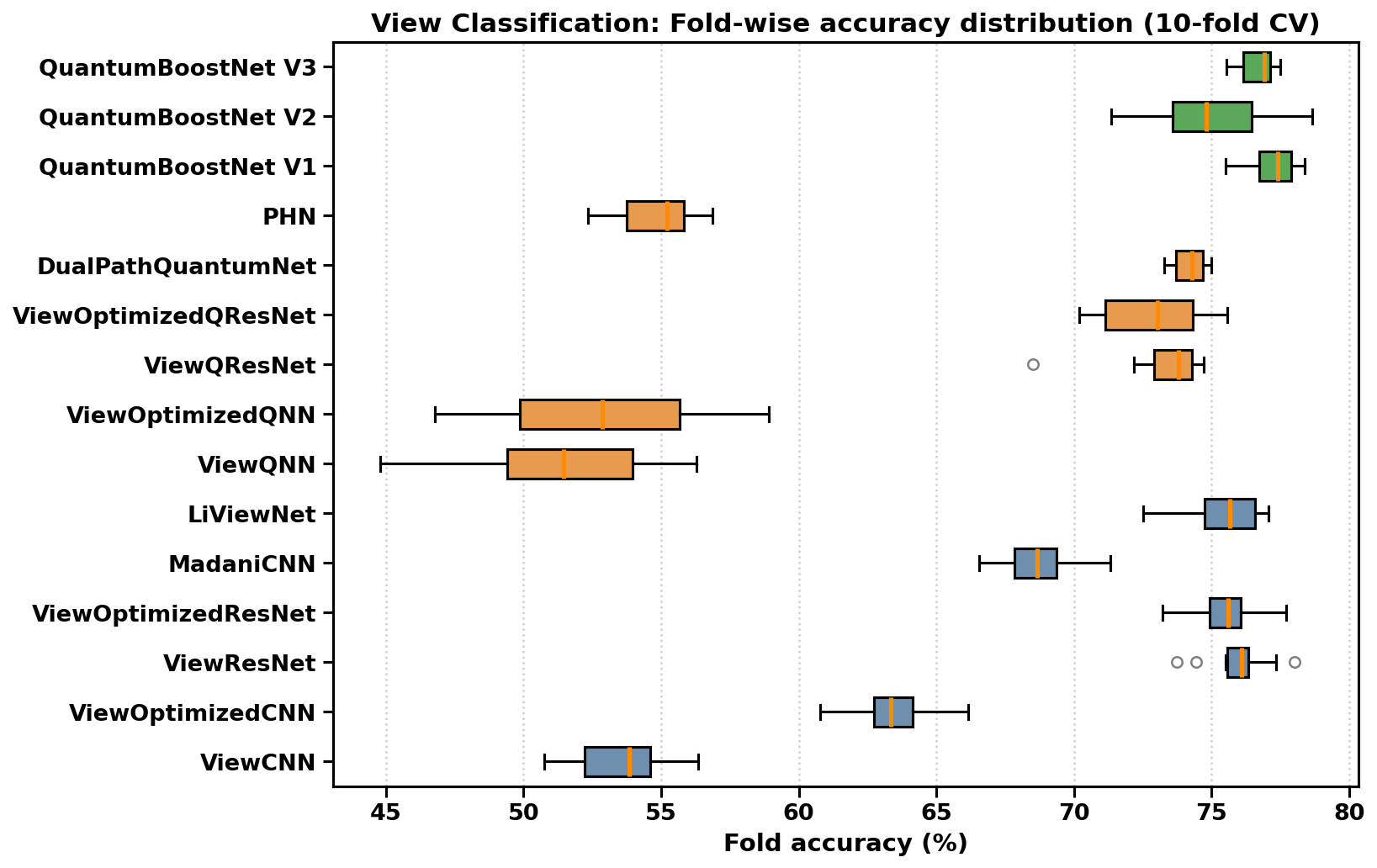}
\caption{Fold-wise accuracy distributions on the View Classification dataset (10-fold CV) in S1. Blue: classical; orange: hybrid; green: \emph{QuantumBoostNet}.}
\label{fig:vc_boxplot}
\end{figure}

Fig.~\ref{fig:vc_boxplot} shows the fold-wise distribution of test accuracies for all models on the View Classification dataset in S1. The narrow interquartile ranges of \emph{QuantumBoostNet V1} and \emph{V3} confirm their consistency across folds, whereas models such as \emph{ViewOptimizedQNN} and \emph{ViewQNN} exhibit high variance. 

\begin{table}[!t]
\centering
\caption{Average per-class F1 score (\%) (10-fold CV) on the View Classification dataset for S1. Best F1 score per class in bold.}
\label{tab:per_class_f1}
\setlength{\tabcolsep}{4pt}
\renewcommand{\arraystretch}{1.05}
\resizebox{\columnwidth}{!}{%
\begin{tabular}{lcccccccccccccc}
\toprule
\textbf{Model} & \textbf{02} & \textbf{06} & \textbf{08} & \textbf{12} & \textbf{13} & \textbf{14} & \textbf{15} & \textbf{16} & \textbf{18} & \textbf{21} & \textbf{22} & \textbf{23} & \textbf{27} & \textbf{30} \\
\midrule
\emph{ViewCNN} & 77.0 & 2.9 & 47.1 & 47.8 & 40.7 & 0.9 & 67.5 & 38.7 & 3.8 & 49.2 & 23.1 & 55.4 & 71.7 & 59.2 \\
\emph{ViewOptimizedCNN} & 82.9 & 7.3 & 63.3 & 56.1 & 44.5 & 8.9 & 76.6 & 60.7 & 22.2 & 62.1 & 18.0 & 69.4 & 87.5 & 72.4 \\
\emph{ViewResNet} & 92.8 & 39.3 & 83.2 & 63.4 & 51.6 & 17.3 & 88.0 & 72.1 & 70.1 & 76.9 & 51.3 & 88.0 & 90.7 & 89.1 \\
\emph{ViewOptimizedResNet} & 92.2 & 48.3 & 81.8 & 63.8 & 49.7 & 18.7 & 87.7 & 70.1 & 66.9 & 78.7 & 52.4 & 87.4 & 90.5 & 91.5 \\
\emph{MadaniCNN} & 86.4 & 1.1 & 73.2 & 52.8 & 44.4 & 6.4 & 84.3 & 63.0 & 37.7 & 78.6 & 53.0 & 79.9 & 87.9 & 83.9 \\
\emph{LiViewNet} & 89.8 & 40.2 & 81.1 & 62.2 & 49.5 & \textbf{19.9} & 88.2 & 73.8 & 67.3 & \textbf{83.1} & \textbf{65.0} & 86.2 & 87.1 & 83.8 \\
\midrule
\emph{ViewQNN} & 72.6 & 1.2 & 47.7 & 47.1 & 40.9 & 4.0 & 67.5 & 27.0 & 8.2 & 49.7 & 22.6 & 51.7 & 60.3 & 54.7 \\
\emph{ViewOptimizedQNN} & 72.8 & 2.1 & 48.5 & 43.9 & 41.8 & 7.7 & 69.4 & 50.5 & 18.2 & 43.6 & 27.3 & 57.1 & 72.9 & 66.5 \\
\emph{ViewQResNet} & 88.5 & 29.8 & 75.7 & 60.2 & 47.7 & 11.4 & 86.0 & 70.3 & 61.9 & 78.2 & 62.5 & 85.4 & 87.6 & 84.1 \\
\emph{ViewOptimizedQResNet} & 90.0 & 16.4 & 79.4 & 60.9 & 47.3 & 11.1 & 86.5 & 71.7 & 57.7 & 77.3 & 59.8 & 85.5 & 84.3 & 79.5 \\
\emph{DualPathQuantumNet} & 90.0 & 23.1 & \textbf{88.4} & \textbf{68.0} & \textbf{58.1} & 3.6 & 85.4 & 57.7 & 24.4 & 75.8 & 18.5 & 76.8 & 86.7 & 82.3 \\
\emph{PHN} & 76.0 & 5.0 & 52.7 & 47.4 & 43.1 & 4.5 & 70.8 & 41.1 & 14.7 & 50.2 & 28.7 & 54.9 & 72.6 & 61.6 \\
\midrule
\emph{QuantumBoostNet V1} & \textbf{93.6} & \textbf{51.3} & 84.8 & 63.2 & 49.9 & 14.7 & \textbf{89.7} & \textbf{74.4} & \textbf{79.2} & 79.4 & 53.1 & \textbf{89.3} & 91.0 & 90.9 \\
\emph{QuantumBoostNet V2} & 89.9 & 33.6 & 79.9 & 62.5 & 49.5 & 16.0 & 87.1 & 72.6 & 57.8 & 81.5 & 62.3 & 86.7 & 89.8 & 87.8 \\
\emph{QuantumBoostNet V3} & 92.8 & 36.8 & 82.9 & 63.9 & 50.1 & 16.8 & 88.2 & 71.3 & 65.7 & 82.7 & 58.0 & 88.6 & \textbf{92.0} & \textbf{92.1} \\
\bottomrule
\end{tabular}%
}
\end{table}

Tables~\ref{tab:vc_clean} and ~\ref{tab:per_class_f1} showcase the macro scores for our models. While \emph{QuantumBoostNet V1} achieves the highest accuracy ($77.19\% \pm 0.89\%$), macro recall (70.8\%), macro F1 (71.8\%), weighted precision (76.8\%), and weighted F1 (76.7\%), \emph{DualPathQuantumNet} attains the highest macro precision (78.3\%) despite a lower accuracy of $74.21\% \pm 0.57\%$ and lower macro recall (58.9\%) and macro F1 (59.8\%). The divergence reflects different operating points on the precision--recall trade-off across the 14 classes: macro-averaged metrics weight every class equally, so a conservative precision behaviour on rare classes inflates macro precision without proportionally affecting accuracy, whereas support-weighted metrics and accuracy are dominated by high-frequency classes. Consistent with this, \emph{QuantumBoostNet V1} and \emph{V3} attain the two highest weighted precisions (76.8\% and 76.3\%), indicating that the \emph{QuantumBoostNet} family excels on the dominant clinical views.


\subsubsection{Results for setting S2}

Table~\ref{tab:noisy} reports the test accuracy of the hybrid models under noisy conditions, both simulated and on real hardware from IBM (more specifically, the Heron r2 device \emph{ibm\_marrakesh}, with \texttt{resilience\_level} being set to 1). All were applied to the View Classification dataset (see Section~IV-C). These results showcase the resistance of our \emph{QuantumBoostNet} models to noisy environments.

\begin{table}[!t]
\caption{Test accuracy (\%) of the QuantumBoostNet models under noisy conditions on the View Classification dataset in S2. Best result per column in bold.}
\label{tab:noisy}
\centering
\footnotesize
\begin{tabular}{l c c  c}
\toprule
Model & Dep. & AD & Avg. \\
\midrule
\emph{QuantumBoostNet V1 (Simulated NISQ)} &         77.01  & \textbf{78.25} &         77.63 \\
\emph{QuantumBoostNet V2 (Simulated NISQ)} &         74.49  &         72.45  &         73.47 \\
\emph{QuantumBoostNet V3 (Simulated NISQ)} & \textbf{78.18} &         77.32  & 77.75 \\
\emph{QuantumBoostNet V1 (QPU)} & - & - & \textbf{77.78} \\
\bottomrule
\end{tabular}
\end{table}



\subsubsection{Results for settings S3 and S4}

To verify that \emph{QuantumBoostNet's} advantage is not dataset-specific, all 15 architectures were also evaluated on FashionMNIST~\cite{xiao2017fashion} in S3 and on MNIST~\cite{lecun1998gradient} in S4 under the same 10-fold cross-validation protocol. Table~\ref{tab:benchmarks} summarizes the results.

\begin{table}[!t]
\caption{Mean accuracy (\%) $\pm$ SD on FashionMNIST in S3 and MNIST in S4 (10-fold CV). Best result per column in bold.}
\label{tab:benchmarks}
\centering
\footnotesize
\begin{tabular}{l c c}
\toprule
Model & FashionMNIST & MNIST \\
\midrule
\emph{ViewCNN}              & 88.09 $\pm$ 0.19 & 98.42 $\pm$ 0.11 \\
\emph{ViewOptimizedCNN}     & 92.25 $\pm$ 0.15 & 99.30 $\pm$ 0.06 \\
\emph{ViewResNet}           & 93.03 $\pm$ 0.23 & 99.36 $\pm$ 0.04 \\
\emph{ViewOptimizedResNet}  & 93.08 $\pm$ 0.23 & 99.41 $\pm$ 0.04 \\
\emph{MadaniCNN}            & 92.27 $\pm$ 0.27 & 99.44 $\pm$ 0.05 \\
\emph{LiViewNet}            & 92.15 $\pm$ 0.23 & 99.43 $\pm$ 0.02 \\
\midrule
\emph{ViewQNN}              & 88.01 $\pm$ 0.54 & 98.07 $\pm$ 0.21 \\
\emph{ViewOptimizedQNN}     & 91.62 $\pm$ 0.37 & 99.15 $\pm$ 0.10 \\
\emph{ViewQResNet}          & 91.80 $\pm$ 0.74 & 99.28 $\pm$ 0.14 \\
\emph{ViewOptimizedQResNet} & 91.39 $\pm$ 1.35 & 99.22 $\pm$ 0.10 \\
\emph{DualPathQuantumNet}   & 90.23 $\pm$ 0.17 & 98.04 $\pm$ 0.13 \\
\emph{PHN}                  & 89.31 $\pm$ 0.36 & 98.73 $\pm$ 0.12 \\
\midrule
\emph{QuantumBoostNet V1}   & 93.55 $\pm$ 0.27 & 99.56 $\pm$ 0.05 \\
\emph{QuantumBoostNet V2}   & 92.71 $\pm$ 0.42 & 99.36 $\pm$ 0.08 \\
\emph{QuantumBoostNet V3}   & \textbf{93.96 $\pm$ 0.24} & \textbf{99.61 $\pm$ 0.04} \\
\bottomrule
\end{tabular}
\end{table}

On MNIST, \emph{QuantumBoostNet~V3} achieves the highest accuracy of 99.61 $\pm$ 0.04\%, outperforming all classical models. On FashionMNIST, \emph{QuantumBoostNet~V3} again ranks first at 93.96 $\pm$ 0.24\%. Across the settings S1, S3, and S4, the \emph{QuantumBoostNet} family consistently occupies the top positions. Table~\ref{tab:vc_clean} and Table~\ref{tab:benchmarks} confirm the global dominance of the \emph{QuantumBoostNet} variants and reveal the architecture-dependence of the quantum advantage: simple hybrid models (\emph{ViewQNN}, \emph{ViewOptimizedQNN}) consistently rank among the weakest across all datasets, whereas ResNet-backed hybrids and the \emph{QuantumBoostNet} family remain competitive.


\subsubsection{Statistical analysis}

To assess whether the observed performance differences are statistically meaningful, we conducted formal hypothesis tests on the fold-level accuracy vectors. For each dataset, the best classical model and the best \emph{QuantumBoostNet} variant--evaluated on the same 10 cross-validation folds--were compared using two complementary tests: the paired Student $t$-test and the Wilcoxon signed-rank test. Effect sizes are reported as Cohen's $d$ on the paired differences. The significance level is set at $0.05$, see Table~\ref{tab:stat_tests}.

\begin{table*}[!t]
\caption{Paired statistical comparison between the best classical model and the best \emph{QuantumBoostNet} variant per dataset. Positive $\Delta$ indicates that the \emph{QuantumBoostNet} model outperforms the classical baseline.}
\label{tab:stat_tests}
\centering
\footnotesize
\begin{tabular}{l l l c c c c c c c}
\toprule
Dataset & Best Classical & Best \emph{QuantumBoostNet} & $n$ & $\Delta$ (pp) & $t$ & $p_t$ & $W$ & $p_W$ & Cohen's $d$ \\
\midrule
View Classification & \emph{ViewResNet}          & \emph{QuantumBoostNet V1} & 10 & $+1.24$ & 2.05 & 0.0711 & 11.0 & 0.1055 & 0.65 \\
FashionMNIST        & \emph{ViewOptimizedResNet} & \emph{QuantumBoostNet V3} & 10 & $+0.89$ & 8.84 & $9.91\!\times\!10^{-6}$ & 0.0 & 0.0020 & 2.79 \\
MNIST               & \emph{MadaniCNN}           & \emph{QuantumBoostNet V3} & 10 & $+0.18$ & 8.56 & $1.29\!\times\!10^{-5}$ & 0.0 & 0.0020 & 2.71 \\
\bottomrule
\end{tabular}
\end{table*}

On MNIST, the mean paired advantage of \emph{QuantumBoostNet~V3} over \emph{MadaniCNN} is +0.18\,pp. Both the paired $t$-test ($p_t = 1.29 \times 10^{-5}$) and the Wilcoxon signed-rank test ($p_W = 0.0020$) indicate a highly significant difference. The paired Cohen's $d = 2.71$ represents a very large effect; the Wilcoxon statistic $W = 0$ indicates \emph{QuantumBoostNet~V3} outperformed the classical baseline on every fold.

On FashionMNIST, the advantage is larger in absolute terms. \emph{QuantumBoostNet~V3} exceeds \emph{ViewOptimizedResNet} by +0.89\,pp on average. Both tests again confirm significance ($p_t = 9.91 \times 10^{-6}$, $p_W = 0.0020$), with Cohen's $d = 2.79$.

On the View Classification dataset, the mean improvement of \emph{QuantumBoostNet~V1} over \emph{ViewResNet} is +1.24\,pp. This difference does not reach statistical significance at $0.05$ under either test ($p_t = 0.071$, $p_W = 0.106$), and the effect size is medium ($d = 0.65$). The result is consistent with the higher fold-to-fold variance on this more challenging task. However, the $p$-values are close to the significance boundary, suggesting that modest increases in statistical power, through additional cross-validation folds, repeated random seeds, or variance-reduction strategies, could resolve this question.

To assess whether model identity has a significant overall effect on fold accuracy, a Friedman test~\cite{friedman1937use} was performed on each dataset, treating the 15 models as treatments and the 10 folds as blocks. The results are reported in Table~\ref{tab:friedman}.

\begin{table}[!t]
\caption{Friedman omnibus test per dataset. All $p$-values indicate that model identity has a significant effect on fold accuracy.}
\label{tab:friedman}
\centering
\footnotesize
\begin{tabular}{l c c c}
\toprule
Dataset & $k$ & $\chi^2$ & $p$-value \\
\midrule
View Classification & 15 & 124.34 & $8.86\!\times\!10^{-20}$ \\
FashionMNIST        & 15 & 127.89 & $1.77\!\times\!10^{-20}$ \\
MNIST               & 15 & 127.17 & $2.45\!\times\!10^{-20}$ \\
\bottomrule
\end{tabular}
\end{table}

In all datasets, the Friedman test yields $p$-value $<$ $10^{-18}$, rejecting the null hypothesis that all models perform equally. This confirms that the performance differences reported in Tables~\ref{tab:vc_clean}--\ref{tab:benchmarks} are not attributable to random fold-to-fold variation.

To summarize the global performance ordering, the average rank of each model was computed across all three datasets and all 30 folds (10 per dataset).
Within each (dataset, fold) pair, models were ranked by accuracy, and the mean rank across all 30 evaluations was taken. The five highest-ranked models are:
(1) \emph{QuantumBoostNet~V3} with a mean rank of 1.70,
(2) \emph{QuantumBoostNet~V1} at 2.00,
(3) \emph{ViewOptimizedResNet} at 4.43,
(4) \emph{ViewResNet} at 4.88, and
(5) \emph{QuantumBoostNet~V2} at 5.47.
The \emph{QuantumBoostNet} family occupies the top two and the fifth position out of 15 models.

\subsubsection{Ablation study} To isolate the contribution of the quantum head from that of the two-phase training protocol, we constructed a two-classical-head control, \emph{QuantumBoostNet V1-CA}. The control is identical to \emph{QuantumBoostNet V1} in every component except one. The variational quantum circuit is replaced by a classical block with an exactly matched parameter count (120, equal to the circuit's trainable angles), operating on the same 10-dimensional bottleneck with outputs bounded in $[-1,1]$ to match the Pauli-$Z$ expectation range. Otherwise, the model remains unchanged. The control was evaluated under the same 10-fold protocol as S1. The V1 variant was chosen for its consistently high performance on the View Classification dataset. \emph{V1-CA} has an average accuracy of $76.80\%$, suggesting that the quantum variant does hold an advantage over the ablation.


\section{Conclusions}

The QuantumBoostNet family is among the best-performing systems on all the benchmarks when a principled training protocol is used. The two-phase approach achieves the highest mean accuracy of any of the models on all three benchmarks. The best-performing competing model reaches an accuracy of 75.95\%, whereas QuantumBoostNet attains 77.19\%; this corresponds to a relative improvement of 1.63\% in the identification of cardiac ultrasound views. This level of improvement is statistically significant on both MNIST and FashionMNIST ($p_W$ = 0.002, Cohen's $d$ $>$ 2.7). For the more difficult view-classification task, QuantumBoostNet~V1 obtains a higher mean accuracy (an increase of +1.24 percentage points), although the difference does not achieve statistical significance ($p_t$ = 0.071). These results provide empirical support for the further development of structured hybrid training strategies in medical image analysis. Effective hybridization requires a quality backbone. The findings indicate the main limitation in hybrid classification is the expressive power of the classical feature extractor. Quantum heads work best when placed on strong pre-trained or co-trained representations.

The two-phase training protocol enables effective quantum integration. The architecture of the quantum circuit alone does not determine classification performance; the training protocol is equally important. The \emph{QuantumBoostNet} framework uses the same 10-qubit circuit but divides training into two phases so each branch can specialise before combining. The switch between phases is dynamic. This method reduces gradient interference and co-adaptation problems that occur when classical and quantum parameters are optimised simultaneously. It also includes a noise-robust fallback: if the quantum branch is affected by gate errors, the classical head retains a well-trained representation that stabilises the final prediction.

An advantage of this study is that the quantum circuits were not only simulated, but tested on real hardware as well, with the models showcasing robustness to noise. Two promising future research areas are increasing the variational circuit size beyond 10 qubits by lowering the feature-to-qubit compression ratio and applying the two-phase training protocol to other medical imaging tasks where noise robustness is important.

\section*{Acknowledgement}

This work was supported by the EMAH Stiftung Karla VÖLLM, Krefeld, Germany, and by RExQTCS: Romanian Excellence Quantum Technologies Enhancing Cybersecurity, a grant of the Ministry of Education and Research, CCCDI - UEFISCDI, project number PN-IV-P6-6.1-CoEx-2024-0214, within PNCDI IV. The authors wish to thank FreeYa Mind Campus for enabling the experiment on real quantum hardware by providing access to IBM’s quantum processing units.

\end{document}